\documentclass[11pt]{article}

\usepackage[preprint]{acl}

\usepackage{times}
\usepackage{latexsym}
\usepackage{booktabs}
\usepackage{graphicx}
\usepackage[table]{xcolor}
\usepackage{longtable}
\usepackage{array}
\usepackage{caption}

\usepackage{listings}
\usepackage[most]{tcolorbox}
\usepackage{hyperref}

\newcounter{promptctr}
\renewcommand{\thepromptctr}{Prompt~\Alph{promptctr}}

\definecolor{promptbg}{HTML}{F6F6F2}
\definecolor{prompttitle}{HTML}{E8F4FF}

\tcbset{
  promptstyle/.style={
    colback=promptbg,
    colframe=black!25,
    coltitle=black,
    fonttitle=\bfseries,
    colbacktitle=prompttitle,
    boxrule=0.4pt,
    arc=2pt,
    left=4pt,right=4pt,top=4pt,bottom=4pt,
    toptitle=2pt,bottomtitle=2pt,
    breakable
  }
}

\newcommand{\promptboxwide}[3]{%
  \refstepcounter{promptctr}
  \begin{tcolorbox}[title={\thepromptctr: #2},promptstyle,width=\linewidth]
    \lstinputlisting{#1}
    \label{#3}
  \end{tcolorbox}
}

\definecolor{bestgreen}{RGB}{226,239,218}
\definecolor{bestgray}{RGB}{242,242,242}

\usepackage[T1]{fontenc}

\usepackage[utf8]{inputenc}
\usepackage{amsmath}

\usepackage{microtype}

\usepackage{inconsolata}

\usepackage{graphicx}
\usepackage{pifont}
\title{
\textsc{SCoPE}: Planning for Hybrid Querying over Clinical Trial Data}

\author{
\textsuperscript{1}Suparno Roy Chowdhury\thanks{These authors contributed equally.} \quad
\textsuperscript{1}Manan Roy Choudhury\footnotemark[1] \quad
\textsuperscript{1}Tejas Anvekar \\
\textbf{\textsuperscript{2}Muhammad Ali Khan} \quad
\textbf{\textsuperscript{2}Kaneez Zahra Rubab Khakwani} \quad
\textbf{\textsuperscript{2}Mohamad Bassam Sonbol} \\
\textbf{\textsuperscript{2}Irbaz Bin Riaz}\thanks{Corresponding authors.} \quad
\textbf{\textsuperscript{1}Vivek Gupta}\footnotemark[2] \\
\textsuperscript{1}Arizona State University \quad
\textsuperscript{2}Mayo Clinic \\
\texttt{\{srchowd3,mroycho1,tanvekar,vgupt140\}@asu.edu} \\
\texttt{\{khan.muhammad2,Khakwani.kaneezzahra,Sonbol.mohamad,riaz.irbaz\}@mayo.edu}
}

\begin{document}
\maketitle
\begin{abstract}
We study \emph{clinical trial table reasoning}, where answers are not directly stored in visible cells but must be reasoned from semantic understanding through normalization, classification, extraction, or lightweight domain reasoning. Motivated by the observation that current LLM approaches often suffer from ``bad reasoning'' under implicit planning assumptions, we focus on settings in which the model must recover implicit attributes such as therapy type, added agents, endpoint roles, or follow-up status from partially observed clinical-trial tables. We propose \textsc{SCoPE} (\emph{\textbf{S}tructured \textbf{C}linical hybrid \textbf{P}lanning for \textbf{E}vidence retrieval in clinical trials}), a multi-LLM planner-based framework that decomposes the task into row selection, structured planning, and execution. The planner makes the source field, reasoning rules, and output constraints explicit before answer generation, reducing ambiguity relative to direct prompting. We evaluate \textsc{SCoPE} on 1{,}500 hybrid reasoning questions over oncology clinical-trial tables against zero-shot, few-shot, chain-of-thought, TableGPT2, BlendSQL, and EHRAgent. Results show that explicit multi-LLM planning improves accuracy for reasoning-based questions while offering a stronger accuracy-efficiency tradeoff than heavier agentic baselines. Our findings position clinical trial reasoning as a distinct table understanding problem and highlight hybrid planner-based decomposition as an effective solution.
\end{abstract}

\begin{figure}[t]
    \centering
        \includegraphics[width=0.975\linewidth]{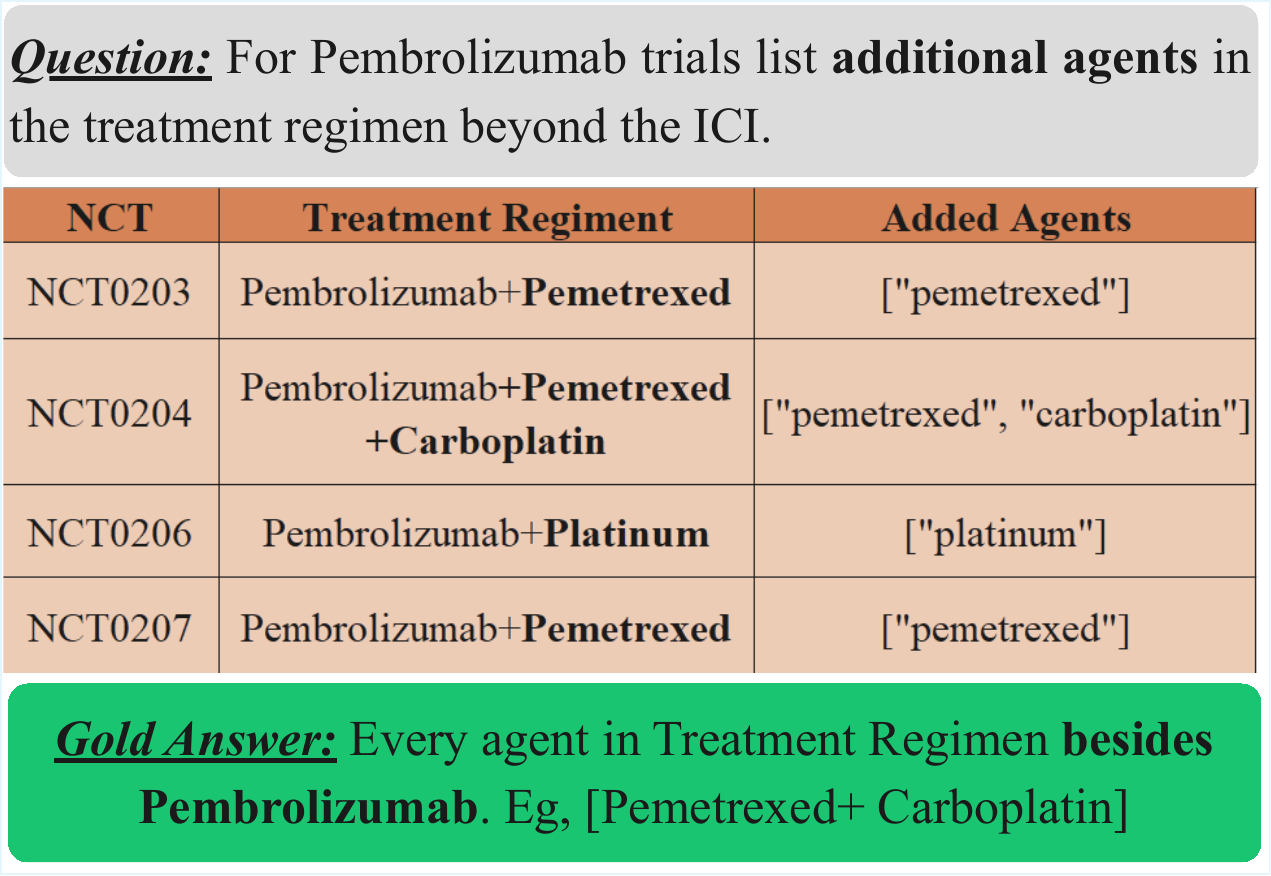}
    \caption{Sample exemplar hybrid reasoning question requiring semantic understanding of regimen structure: the model must recognize \textbf{Pembrolizumab} as the ICI named in the question and extract all other treatment agents from the visible regimen text as the held-out target field. For instance, in NCT0203, \textbf{Pembrolizumab} is the ICI and \textbf{Pemetrexed} is the additional agent used in the treatment regimen}
    \label{fig:hybrid_reasoning_example}
\end{figure}

\section{Introduction}
\begin{figure*}[t]
    \centering
    \includegraphics[width=1.0\textwidth]{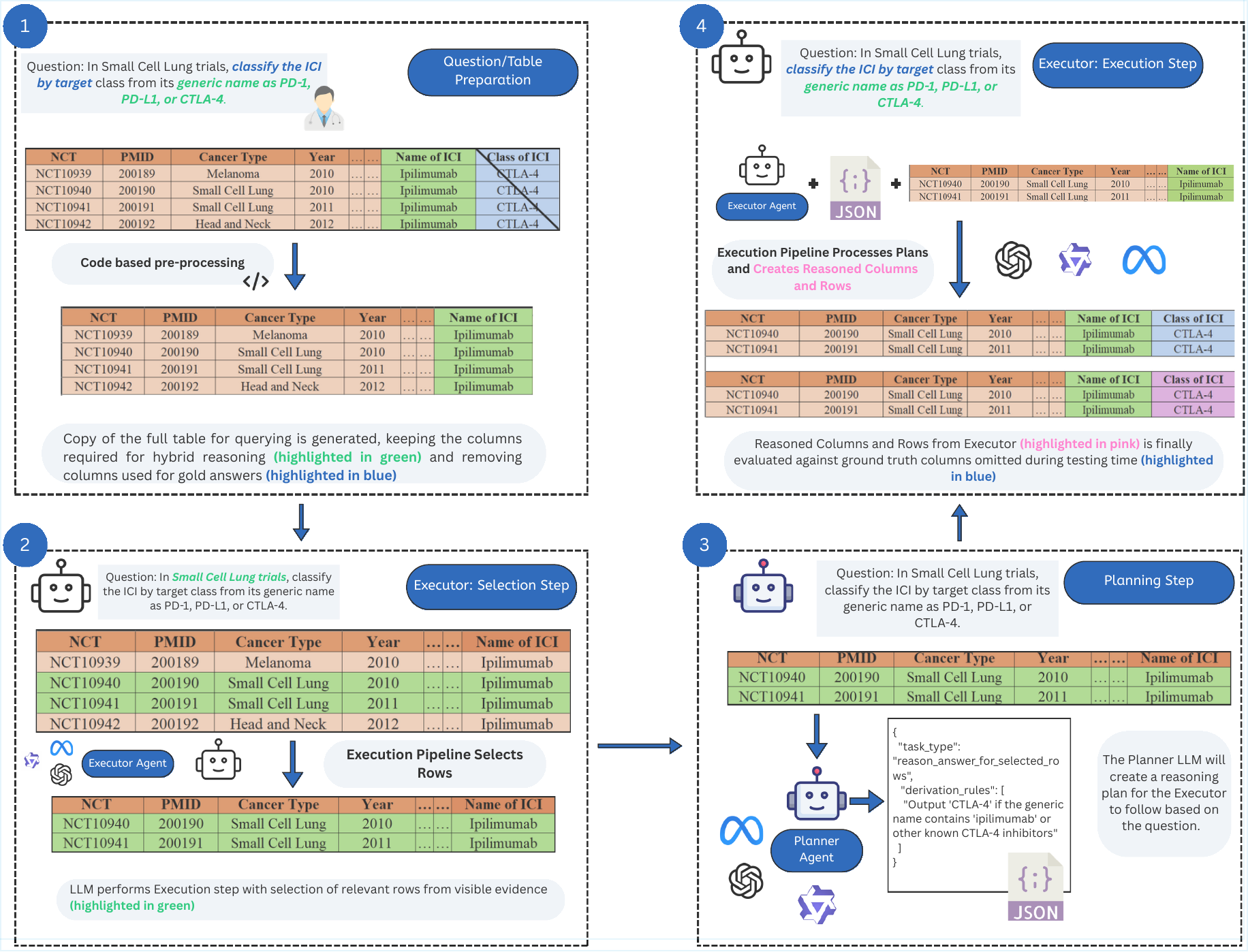}
    \caption{\textsc{\textsc{SCoPE}} for clinical tabular reasoning. Given a question and a copied visible table, \textsc{\textsc{SCoPE}} first prepares an inference-time table view by retaining the visible evidence columns and withholding the target column used for evaluation if there are any. The executor identifies the rows relevant to the question, the planner produces a structured reasoning plan over the selected table, and the executor follows this plan to generate the final row-aligned predictions. In this example, the method reasons over the \emph{Name of ICI} column to recover the held-out \emph{Class of ICI} field, and the resulting predictions are compared against the omitted ground-truth targets.}
    \label{fig:scope_pipeline}
\end{figure*}

Many clinical trial questions are neither pure retrieval nor standard text-to-SQL. In these cases, the target value is not explicitly stored in a visible cell and cannot be obtained by simply selecting rows or projecting an existing column. Instead, it must typically be derived from row contents through normalization, classification, extraction, aggregation, or lightweight domain reasoning. We study this setting as \emph{clinical trial reasoning}: given a question and a partially observed table, the system must identify the relevant rows, ground the visible source field containing the evidence, and apply the correct row-wise transformation to produce a target field that is often absent from the schema. This setting is related to table question answering and semantic parsing, but is only partially captured by existing table QA benchmarks and classical text-to-SQL formulations \cite{pasupat2015compositional, zhong2017seq2sql, yu2018spider, chen2019tabfact, nan2022fetaqa}.

Clinical trial reasoning arises naturally in systematic reviews, where analysts often need to infer implicit attributes from raw fields such as treatment regimens, endpoint descriptions, follow-up text, and bibliographic metadata. Typical questions require medically meaningful outputs such as therapy type, added agents, endpoint role, follow-up maturity, platinum status, or drug class. These questions are often compositional: solving them requires correct row selection, correct source-field grounding, and correct transformation of visible evidence. Errors in any of these steps can lead to incorrect or unusable outputs, making reliable decomposition critical for this setting.

Existing approaches do not fully address this problem. Direct prompting with large language models often under-structures the task, typically relying on a single generation step to retrieve rows, infer the evidence-bearing field, apply the transformation, and format row-aligned answers. Text-to-SQL methods assume that the answer can be recovered through executable operations over explicit schema values, which is often not true when the target field is latent rather than stored \cite{zhong2017seq2sql, yu2018spider}. More structured agentic and program-based approaches can improve reasoning, but can introduce additional execution overhead and brittle multi-step traces \cite{gao2023pal, cheng2022binding, yao2022react, glenn2024blendsql, shi2024ehragent}. So, the main research question we are trying to address is: \emph{Can explicit hybrid planner-based decomposition improve grounded row-level reasoning over partially observed clinical-trial tables, and do the resulting gains appear more clearly in row-level grounding than in aggregate answer recovery?}

To address these limitations, we introduce \textsc{SCoPE}, a planner-based framework for clinical querying that explicitly decomposes the task into row selection, structured planning, and execution. The executor grounds the question to relevant rows in the visible table and produces row-aligned outputs, while the planner interprets the question as a structured transformation over the selected evidence. Concretely, the executor first identifies a candidate subset of rows, the planner converts the question into an explicit transformation specification, and the executor then applies that specification to generate aligned predictions. This decomposition makes the source-field choice, reasoning rules, and output constraints explicit prior to answer generation, improving interpretability and reducing ambiguity compared to direct prompting. By introducing an explicit planning interface between reasoning and execution, \textsc{\textsc{SCoPE}} better aligns the model behavior with the compositional structure of clinical trial querying.

We also construct a benchmark of 1{,}500 hybrid reasoning questions over oncology clinical-trial tables, where target attributes must be inferred from visible evidence rather than directly retrieved.

Unlike approaches that keep planning implicit in a single generation step or rely on heavier agentic pipelines, \textsc{SCoPE} introduces a lightweight planning interface that separates reasoning specification from execution while preserving grounding in the visible table. We evaluate \textsc{SCoPE} against zero-shot, few-shot, chain-of-thought, TableGPT2, BlendSQL, and EHRAgent, and find that explicit planning improves reasoning accuracy while offering a better accuracy-efficiency tradeoff than heavier agentic baselines. 

Our main contributions are:
\begin{itemize}
    \item We introduce \textsc{SCoPE}, a planner-based framework that decomposes clinical trial reasoning into row selection, source-field grounding, and structured execution.
    \item We formalize \emph{clinical trial reasoning} as a distinct table understanding setting requiring inference over visible row evidence rather than direct cell retrieval.
    \item We construct a benchmark of 1{,}500 oncology clinical-trial questions and show that explicit planning outperforms prompting baselines while remaining more efficient than heavier agentic methods.
\end{itemize}

\noindent\textbf{Problem formulation.}
Given a question $\tilde{q}$ and a visible table $X = \{r_1, \dots, r_m\}$ with a hidden target field, the goal is to recover a set of row-aligned predictions $\hat{G} = \{(r, \hat{y}_r)\}$ for the subset of rows relevant to the query. Each $\hat{y}_r$ must be derived from visible evidence in $r$ rather than retrieved directly from a stored field. Errors may arise from incorrect row selection, incorrect source-field grounding, or incorrect transformation of visible evidence, making joint reasoning over these steps essential.

\section{Dataset}
\subsection{Dataset Construction}

We begin with a seed set of $500$ questions authored by a certified oncology researcher to reflect realistic evidence-review queries over a structured clinical-trial table containing bibliographic metadata, disease labels, treatment regimens, control-arm descriptions, endpoint text, trial phase, sample size, and follow-up information. Each seed example is paired with an executable SQLite query, which serves as a scaffold for generating new questions and aligned ground truth. We retain the subset of table-solvable questions that require hybrid reasoning rather than simple retrieval, and expand it to $1{,}500$ questions through programmatic augmentation. For each retained pair $(q,s)$, we generate a new pair $(\tilde{q}, \tilde{s}) = T(q,s)$ by applying a single atomic edit to the structured query, such as modifying the projection, simplifying predicates, swapping attribute values, relaxing thresholds, or changing the evidence field while preserving answerability. We keep only read-only variants that execute successfully and return non-empty results, then de-duplicate the resulting query patterns and rewrite each retained structured query into fluent natural language while preserving its exact semantics.

\paragraph{Ground-Truth Construction.}
For each final question $\tilde{q}$, the corresponding query $\tilde{s}$ identifies a set of relevant rows $R=\{r_1,\dots,r_n\}$ together with a visible evidence field $c_{\text{src}}$ used to derive the hidden target. We then apply a deterministic row-level reasoning function $f$ to each row to produce the ground-truth target value $y_i = f(r_i[c_{\text{src}}])$
for every $r_i \in R$. This function implements the intended task semantics for normalization, categorization, boolean interpretation, and structured extraction, such as mapping an ICI mention to its class or brand name, determining endpoint role from endpoint text, bucketing follow-up duration or publication year, or extracting additional agents, schedules, regimen components, and control backbones from treatment text. To form the benchmark input, we expose only the visible table copy while omitting the target values $\{y_i\}_{i=1}^{n}$. The ground truth is stored separately as the row-aligned pairing $G=\{(r_i, y_i)\}_{i=1}^{n}$.

Crucially, each target value is derived from evidence already present in the table rather than authored independently or requiring external domain knowledge at evaluation time. This design ensures that the benchmark evaluates hybrid reasoning over visible evidence: models must infer latent attributes by interpreting existing table content, rather than generating plausible but unsupported answers.

\begin{table}[t]
\centering
\small
\begin{tabular}{lr}
\toprule
\textbf{Clinical-Trial Table Statistic} & \textbf{Value} \\
\midrule
Rows & 159 \\
Columns & 32 \\
Unique trials (\texttt{NCT}) & 105 \\
Cancer types & 19 \\
ICI names & 13 \\
\bottomrule
\end{tabular}
\caption{Statistics of the underlying clinical-trial table used to construct the benchmark. The table covers trial-level oncology evidence, including study metadata, disease labels, treatment and control regimens, endpoint descriptions, and follow-up information, spanning 105 unique trials across 19 cancer types and 13 immune checkpoint inhibitor (ICI) names.}
\label{tab:clinical_table_stats}
\end{table}

\subsection{Dataset Guidelines}

To create the seed set, a certified clinician authored questions based on the IOTOX living evidence table\footnote{\url{https://iotox.living-evidence.com/}}. They were asked to write realistic evidence-review questions while following three guidelines:
\begin{itemize}
    \item Each question had to be grounded in visible table evidence.
    \item The expected answer had to be precise and structured, typically as a JSON value.
    \item The answer had to be inferable from the table itself, without requiring unsupported external knowledge.
\end{itemize}

\subsection{Dataset Statistics}

As shown in Tables~\ref{tab:clinical_table_stats} and~\ref{tab:ground_truth_stats}, the benchmark is built from a structured clinical-trial table with $159$ records and $32$ columns, spanning $105$ unique \texttt{NCT} identifiers, $131$ unique \texttt{PubMed ID}s, $95$ trial names, $19$ cancer types, and $13$ ICI names from 2010 to 2021. Although compact, the table is information-dense: many target outputs are not exposed as ready-made columns, but must be inferred from regimen text, endpoint descriptions, follow-up fields, and bibliographic metadata. At the benchmark level, the dataset contains $1{,}500$ questions with an average length of $21.2$ tokens and covers $31$ distinct target fields. The answer space is also diverse, including string-valued, list-valued, boolean, and null-only outputs, highlighting that the benchmark evaluates hybrid reasoning across multiple target types rather than a single narrow form of table question answering.

\begin{table}[t]
\centering
\small
\begin{tabular}{lr}
\toprule
\textbf{Question Statistic} & \textbf{Value} \\
\midrule
Total questions & 1,500 \\
Mean question length (tokens) & 21.2 \\
Target fields & 31 \\
\midrule
\textbf{Answer Types} & \textbf{Value} \\
\midrule
String & 957 \\
List & 241 \\
Boolean & 224 \\
Null-only & 78 \\
\bottomrule
\end{tabular}
\caption{Question and answer-level benchmark statistics. \emph{Total questions} gives the size of the benchmark; \emph{Mean question length} reports the average number of tokens per natural-language question; \emph{Target fields} gives the number of distinct hidden target types covered by the dataset for reasoning . On the answer side, we report the number of questions whose gold supervision is primarily \emph{string-valued}, \emph{list-valued}, \emph{boolean-valued}, or \emph{null-only}, where null-only questions are those whose held-out target values are absent for all aligned rows.}
\label{tab:ground_truth_stats}
\end{table}

\section{The \textsc{SCoPE} System \& Flow}

Given a question $\tilde{q}$ and a visible table copy
\[
X=\{r_1,\dots,r_m\},
\]
where the answer-bearing target field is omitted but row identifiers and visible evidence columns are preserved, \textsc{SCoPE} uses two LLM components: an \emph{executor} and a \emph{planner}. The executor grounds the question to relevant rows and produces the final row-level outputs, while the planner interprets the selected evidence as a structured reasoning problem. The planner may use the same backbone as the executor or a different one. We provide the row-selection, planning, and final-execution prompts in Appendix~\ref{app:scope_prompts}.

The method has three steps. First, the executor identifies a candidate subset of relevant rows from the full visible table. Second, the planner receives the question and this candidate table and produces an explicit reasoning plan. Third, the executor follows that plan to generate the final row-aligned output table. In short, the planner determines \emph{what} to derive and from \emph{which} visible evidence, while the executor determines \emph{where} that evidence occurs and returns the final answers.

\paragraph{Executor: Row Selection.}
The first step grounds the question $\tilde{q}$ to the visible table $X$. The executor selects a candidate subset of rows
\[
\hat{R} = E_{\mathrm{sel}}(\tilde{q}, X) \subseteq X.
\]
At this stage, the executor is used only to identify relevant rows; it does not attempt to derive the final answer.

\paragraph{Planner: Structured Reasoning.}
The planner then receives the question $\tilde{q}$ and the candidate table $\hat{R}$ selected by the executor, and predicts a structured plan
\[
\pi = P(\tilde{q}, \hat{R}) = \big(c_{\text{src}}, C_{\text{rel}}, \rho, o\big),
\]
where $c_{\text{src}}$ denotes the inferred visible source column that contains the supporting evidence, $C_{\text{rel}}$ denotes the subset of visible columns retained for downstream execution, $\rho$ denotes the reasoning rules needed to transform the source evidence into the target field, and $o$ denotes the output constraints that specify the expected answer format.

\paragraph{Executor: Final Table Generation.}
Given the planner output $\pi$, the executor operates on the focused candidate table
\[
X_{\pi} = \hat{R}[C_{\text{rel}}],
\]
and produces the final row-aligned prediction table
\[
\hat{G} = E_{\mathrm{ans}}(\tilde{q}, \pi, X_{\pi}) = \{(r,\hat{y}_r)\mid r \in \hat{R}\}.
\]
Each predicted value $\hat{y}_r$ is keyed by its row $r$, so the output remains aligned to the rows in the candidate table rather than collapsing the question into a single free-form answer.

\paragraph{Why Planning Helps.}
A key advantage of planning is that it explicitly separates decision points that are otherwise entangled in single-step generation. In particular, SCOPE decomposes (i) row grounding, (ii) source-field identification, and (iii) transformation into distinct stages, each of which can fail independently in direct prompting. By externalizing these decisions into a structured plan, the model avoids implicit assumptions about where evidence resides and how it should be transformed. This leads to more stable row alignment and reduces cascading errors compared to approaches where these steps are implicitly interleaved.

\section{Experimental Setup}

\subsection{Baselines}

We compare \textsc{SCoPE} against three baseline families under the same input setting. Every method receives the same question $\tilde{q}$ and visible table copy $X$, where the target field is omitted at inference time. When a method produces free text, SQL, or other intermediate outputs, we normalize its result programmatically into the shared row-aligned prediction format
\[
\hat{G}=\{(\hat{r}_i,\hat{y}_i)\}_{i=1}^{k},
\]
so that all systems are evaluated against the same held-out ground truth $G$.

The first family consists of zero-shot, few-shot, and chain-of-thought prompting \cite{brown2020language, wei2022chain}, where the model is given the full visible table $X$ and asked to recover the omitted target field directly without an explicit planning stage. The second includes BlendSQL \cite{glenn2024blendsql} and EHRAgent \cite{shi2024ehragent}, which introduce more intermediate tabular structure reasoning through SQL-like compositional reasoning or heavier agentic execution. The third includes TableGPT2 \cite{li2024table}, a dedicated tabular reasoning model that jointly consumes the table and question to predict the hidden target field directly from the table representation.

\subsection{Models and Hyperparameters}

We evaluate three backbone LLMs: \texttt{Qwen3-30B-A3B-Instruct-2507} \cite{yang2025qwen3}, \texttt{gpt-oss-20b} \cite{agarwal2025gpt}, and \texttt{Llama-3.3-70B-Instruct} \cite{grattafiori2024llama}. All generations use deterministic decoding with temperature $0.0$ and top-$p$ $1.0$. Direct prompting baselines are allowed up to $2048$ generated tokens per question. For \textsc{SCoPE}, the planner and executor are chosen from the same model pool, enabling both same-model and cross-model configurations. The row-selection step uses $768$ tokens, the planner $1024$, and the final executor step $2048$, keeping execution lightweight while reserving more capacity for structured planning.

\subsection{Evaluation Metrics}

We evaluate each predicted table against the held-out ground-truth table using three grounded metrics after one-to-one row alignment. \textbf{Table F1} is our primary metric and measures how well a method recovers the correct hidden target values for the correct rows, balancing missed and spurious predictions. \textbf{Grounded Row Jaccard} gives a stricter overlap-based view by measuring the intersection-over-union between predicted and gold row-level outputs. \textbf{Grounded Fowlkes-Mallows} \cite{fowlkes1983method} complements these metrics by summarizing the balance between grounded precision and recall. Because the target field is omitted at inference time and restored only in held-out supervision, all three metrics evaluate recovery of hidden row-level attributes from visible evidence rather than direct copying.

\noindent\textbf{Impact of row selection errors.}
Incorrect row selection directly affects all three metrics: missing relevant rows lowers recall, while selecting spurious rows lowers precision and overlap. As a result, the metrics jointly capture both row grounding and answer correctness.

\section{Results and Discussion}

\begin{table*}[t]
\centering
\small
\setlength{\tabcolsep}{5pt}
\renewcommand{\arraystretch}{1.14}
\begin{tabular}{lccccccccc}
\toprule
& \multicolumn{3}{c}{\textbf{Qwen3}} & \multicolumn{3}{c}{\textbf{Llama-3.3}} & \multicolumn{3}{c}{\textbf{GPT-OSS}} \\
\cmidrule(lr){2-4} \cmidrule(lr){5-7} \cmidrule(lr){8-10}
\textbf{Method} & \textbf{F1 (\%)} & \textbf{RJ (\%)} & \textbf{FM (\%)} & \textbf{F1 (\%)} & \textbf{RJ (\%)} & \textbf{FM (\%)} & \textbf{F1 (\%)} & \textbf{RJ (\%)} & \textbf{FM (\%)} \\
\midrule

\rowcolor{gray!15}
\multicolumn{10}{c}{\textbf{Tabular Reasoning Methods}} \\
BlendSQL & 11.56 & 6.52 & 20.63 & 5.60 & 5.15 & 5.81 & 7.30 & 6.48 & 7.71 \\
EHRAgent & 32.99 & 29.79 & 33.74 & 30.99 & 28.07 & 31.69 & 34.85 & 31.23 & 35.65 \\

\midrule
\rowcolor{gray!15}
\multicolumn{10}{c}{\textbf{Prompting Technique Methods}} \\
Zero Shot & 56.32 & 44.95 & 62.73 & 66.96 & 54.55 & 72.04 & 73.50 & 61.05 & 77.47 \\
CoT & 55.37 & 44.65 & 61.93 & \textbf{\textcolor{green!50!black}{70.87}} & 57.83 & 75.15 & 74.17 & 61.77 & 78.05 \\
Few-Shot & 54.74 & 44.09 & 61.48 & 69.38 & 56.56 & 74.05 & 73.99 & 61.55 & 77.85 \\

\midrule
\rowcolor{gray!15}
\multicolumn{10}{c}{\textbf{Table Model Method}} \\
TableGPT2 & \multicolumn{3}{c}{\textbf{F1:} 44.03} & \multicolumn{3}{c}{\textbf{RJ:} 33.78} & \multicolumn{3}{c}{\textbf{FM:} 50.81} \\

\midrule
\rowcolor{gray!15}
\multicolumn{10}{c}{\textbf{Ours: \textsc{SCoPE}}} \\
\textsc{SCoPE} & \textbf{\textcolor{green!50!black}{63.19}} & \textbf{\textcolor{green!50!black}{52.07}} & \textbf{\textcolor{green!50!black}{69.45}} & \textbf{\textcolor{green!50!black}{70.87}} & \textbf{\textcolor{green!50!black}{60.66}} & \textbf{\textcolor{green!50!black}{76.12}} & \textbf{\textcolor{green!50!black}{74.31}} & \textbf{\textcolor{green!50!black}{62.48}} & \textbf{\textcolor{green!50!black}{78.27}} \\

\bottomrule
\end{tabular}
\caption{Per-model comparison across baselines and same-backbone \textsc{SCoPE}, where each listed model is used as \emph{both} the planner and the executor. We report \textbf{F1} (Table F1), \textbf{RJ} (Row Jaccard), and \textbf{FM} (Fowlkes-Mallows). Higher is better for all metrics. Best values are shown in bold green. TableGPT2 is a table based model which does not have a Qwen3, Llama-3.3, or GPT-OSS backbone so it is tested individually.}
\label{tab:per_model_same_backbone}
\end{table*}

\subsection{Main Results}

Across all backbone families, the most important result is that \textsc{\textsc{SCoPE}} consistently achieves the strongest grounded performance while remaining lightweight relative to heavier structured baselines. The framework is best overall on GPT-OSS and Qwen3, and tied on Table F1 while stronger on grounding metrics for Llama-3.3. The clearest gain appears for Qwen3, where explicit planning substantially improves row-aligned reasoning over direct prompting. A paired intent-level breakdown in Appendix~\ref{app:error_analysis_intent} helps explain these gains: the largest improvements appear on \emph{Schema / structural labeling} (+4.93 $\Delta$F1, +7.94 $\Delta$RJ), with further gains on \emph{Normalization / classification} (+1.36, +2.30) and \emph{Structured extraction} (+0.84, +2.08).

\paragraph{\textbf{RQ1: Does explicit planning improve performance over direct prompting and tabular baselines?}}
Yes. Table~\ref{tab:per_model_same_backbone} shows that \textsc{\textsc{SCoPE}} is strongest or tied for strongest across all three model families on the grounded metrics. It consistently outperforms BlendSQL, EHRAgent, and TableGPT2, and also improves over the strongest prompting baselines in most settings. This supports the central claim of the paper: explicit planning is more effective than leaving row grounding and reasoning implicit inside a single generation step.

\paragraph{\textbf{RQ2: Are these gains consistent across backbone models?}}
Largely yes, although the size of the gain varies by model. For GPT-OSS, \textsc{\textsc{SCoPE}} improves over the strongest prompting baseline from $74.17$ to $74.31$ Table F1 while also yielding the best grounded overlap scores. For Llama-3.3, it matches the best prompting Table F1 ($70.87$) but improves Row Jaccard and Fowlkes-Mallows, indicating better row alignment. For Qwen3, the gains are largest, showing that the framework generalizes across backbones while being especially useful for weaker implicit reasoners. This pattern is also consistent with the appendix analysis: gains are strongest on schema-grounding-heavy categories, while \emph{Bucketing / discretization} and \emph{Metadata normalization / extraction} remain weak spots at -0.81/-0.90 $\Delta$F1, respectively; see Appendix~\ref{app:error_analysis_intent}.

\paragraph{\textbf{RQ3: Which models benefit most from explicit planning?}}
Qwen3 benefits the most. Its Table F1 increases from $56.32$ under the best prompting baseline to $63.19$ with \textsc{\textsc{SCoPE}}, with corresponding improvements in Row Jaccard and Fowlkes-Mallows. This suggests that explicit planning is particularly helpful when the underlying model is less reliable at performing row grounding and structured reasoning implicitly.

\paragraph\noindent\textbf{Note.} In addition to Table F1, grounded Row Jaccard (RJ) provides a complementary view of performance by measuring intersection-over-union over row-aligned outputs, directly capturing the joint correctness of row selection and target recovery. This is particularly relevant in our setting, where answers must be derived for the correct subset of rows rather than retrieved from explicit cells. Empirically, RJ highlights improvements in grounding quality: for Llama-3.3, \textsc{SCoPE} matches the best Table F1 (70.87) while improving RJ from 57.83 to 60.66; for Qwen3, gains are +6.87 in F1 (56.32 $\rightarrow$ 63.19) alongside a larger +7.12 in RJ (44.95 $\rightarrow$ 52.07); and for GPT-OSS, near-identical F1 (74.17 $\rightarrow$ 74.31) still corresponds to a consistent increase in RJ (61.77 $\rightarrow$ 62.48). These patterns align with the appendix error analysis, where several of the strongest intent categories show larger gains in $\Delta$RJ than in $\Delta$F1; see Appendix~\ref{app:error_analysis_intent}. RJ is therefore a sensitive indicator of grounded row-level alignment, complementing Table F1 in assessing hybrid reasoning performance.


\subsection{Cross-Model Ablation}

We vary the planner and executor backbones to test whether planning and execution benefit from different model strengths. In particular, this ablation asks whether a stronger planner can improve performance even when the execution model is held fixed, and whether these two roles are better treated as complementary rather than identical.

\begin{table}[!t]
\centering
\small
\setlength{\tabcolsep}{6pt}
\renewcommand{\arraystretch}{1.12}
\begin{tabular}{llccc}
\toprule
\textbf{Executor} & \textbf{Planner} & \textbf{F1}(\%) & \textbf{RJ}(\%) & \textbf{FM}(\%) \\
\midrule
GPT-OSS & Qwen3 & 75.07 & 63.74 & 79.26 \\
Qwen3 & GPT-OSS & 59.59 & 48.26 & 66.32 \\
Qwen3 & Llama-3.3 & 62.47 & 51.40 & 68.88 \\
GPT-OSS & Llama-3.3 & \textbf{\textcolor{green!50!black}{75.12}} & \textbf{\textcolor{green!50!black}{63.88}} & \textbf{\textcolor{green!50!black}{79.28}} \\
Llama-3.3 & GPT-OSS & 68.01 & 57.37 & 73.64 \\
Llama-3.3 & Qwen3 & 71.43 & 61.33 & 76.64 \\
\bottomrule
\end{tabular}
\caption{Cross-model ablation of \textsc{SCoPE}. Rows report configurations where the executor and planner use different backbones. Higher is better. The best cross-model result in each metric is shown in bold green.}
\label{tab:cross_model_ablation_vertical_split}
\end{table}

Table~\ref{tab:cross_model_ablation_vertical_split} shows that changing the planner can indeed improve performance. The clearest example is GPT-OSS: when used as both planner and executor in Table~\ref{tab:per_model_same_backbone}, \textsc{\textsc{SCoPE}} reaches $74.31$ Table F1, but replacing the planner with Llama-3.3 increases performance to $75.12$, with similar gains in Row Jaccard and Fowlkes-Mallows. This suggests that the planner benefits from a stronger reasoning model even when execution remains fixed.

Another example appears in the Llama-3.3 executor family. In the same-backbone setting, Llama-3.3 reaches $70.87$ Table F1, but replacing the planner with Qwen3 improves performance to $71.43$, with corresponding gains in Row Jaccard and Fowlkes-Mallows. Together with the GPT-OSS results, this suggests that changing the planner can improve performance even when the executor is held fixed.

\begin{table}[t]
\centering
\small
\setlength{\tabcolsep}{6pt}
\renewcommand{\arraystretch}{1.12}
\begin{tabular}{lccc}
\toprule
\textbf{Coder Model} & \textbf{F1}(\%) & \textbf{RJ}(\%) & \textbf{FM}(\%) \\
\midrule
GPT-OSS & 14.56 & 11.67 & 23.39 \\
Qwen3 & 48.92 & 37.53 & 56.01 \\
Llama-3.3 & \textbf{\textcolor{green!50!black}{63.40}} & \textbf{\textcolor{green!50!black}{51.55}} & \textbf{\textcolor{green!50!black}{69.10}} \\
\bottomrule
\end{tabular}
\caption{Results for the planner-coder baseline across coder backbones. The same models are used for the planning and code generation execution stage. Best results are shown in bold green.}
\label{tab:planner_coder_results}
\end{table}

\subsection{Planner-Coder Baseline}

We include the planner-coder setting as an ablation of whether execution should be made more explicit through programmatic synthesis. The idea is that, if the planner already produces a structured reasoning trace, then compiling that trace into deterministic Python code could make execution more transparent and reproducible. This tests an alternative to grounded LLM execution: executing the plan by first translating it into code.

Table~\ref{tab:planner_coder_results} shows that this formulation performs substantially worse than \textsc{SCoPE} across all backbones. The drop is especially large for GPT-OSS (from $74.21$ to $14.56$ Table F1) and Qwen3 (from $63.19$ to $48.92$), while even the best planner-coder result with Llama-3.3 ($63.40$) remains below the strongest same-backbone \textsc{SCoPE} systems. This suggests that code-generation-based execution is more brittle: the model must infer the procedure, translate it into executable code, and ensure that the code still matches the table context and output format. Overall, hybrid clinical table reasoning appears to benefit more from constrained grounded execution than from open-ended code synthesis.

\section{Model Cost Effectiveness Analysis}

\begin{figure}[t]
    \centering
    \includegraphics[width=1.0\linewidth]{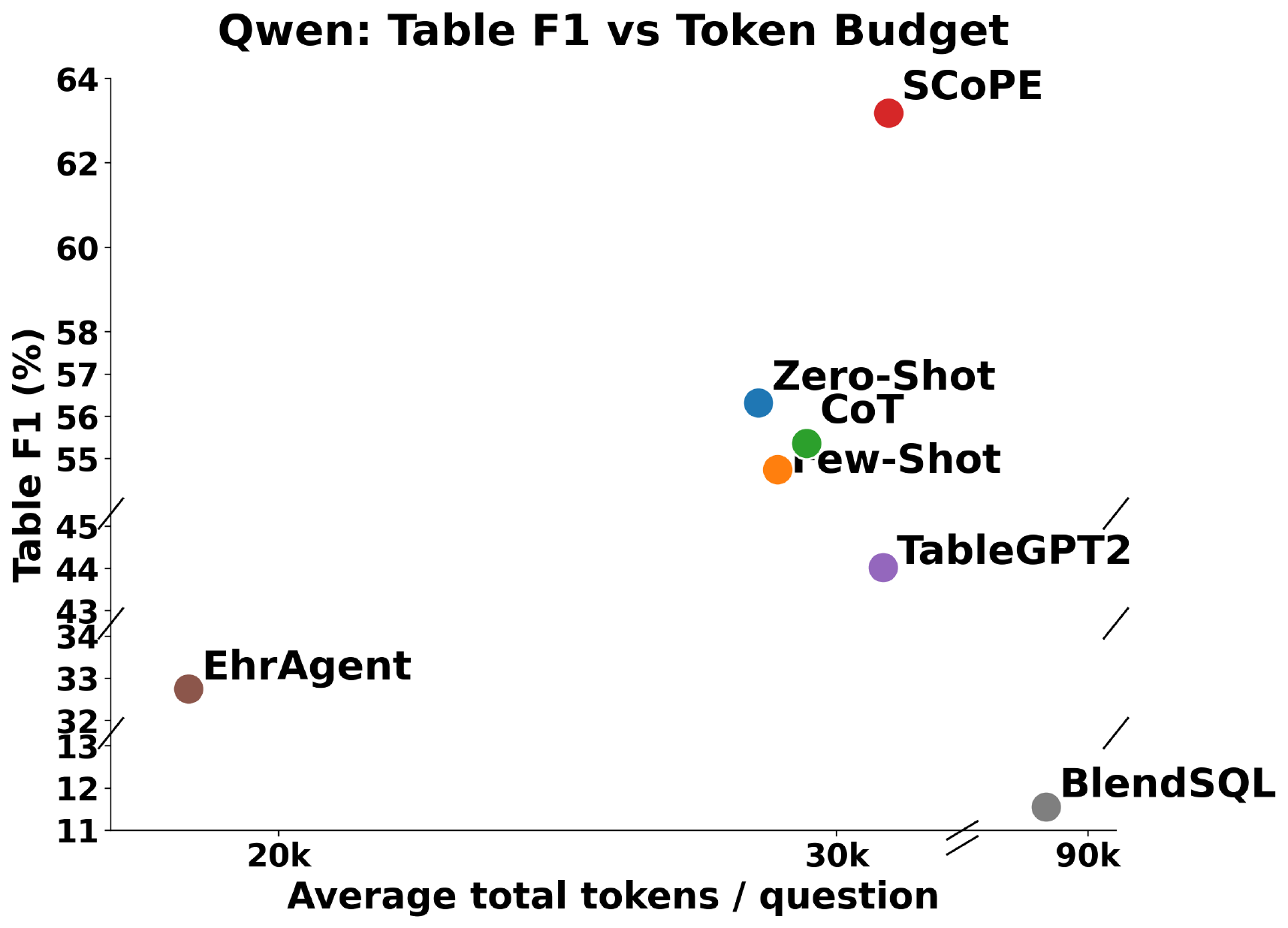}
    \caption{Cost-effectiveness comparison for Qwen-based methods. The x-axis shows average total tokens per question and the y-axis shows \textbf{Table F1}.}
    \label{fig:qwen_cost_effectiveness}
\end{figure}

Figure~\ref{fig:qwen_cost_effectiveness} compares answer quality against token budget in the Qwen-based setting. Although TableGPT2 is not Qwen-based, we include it as a reference point for cost-effectiveness against a dedicated table model. \textsc{SCoPE} lies on the strongest accuracy-cost frontier, achieving the highest Table F1 at roughly $63\%$ with only a modestly larger token budget than direct prompting baselines. Zero-shot, few-shot, and chain-of-thought prompting cluster in the $28$k-$30$k token range but remain in the mid-$50\%$ F1 range, while EHRAgent is cheaper but much less accurate and BlendSQL is both the most expensive and the weakest. Overall, \textsc{SCoPE} adds limited planning overhead while yielding the best accuracy, making it the most favorable cost-effective choice in this comparison.

\section{Related Work}

We situate our work at the intersection of clinical datasets and table reasoning methods, focusing on settings where answers must be inferred rather than directly retrieved.

\paragraph{Related Datasets.}
Clinical datasets such as MIMIC III \& IV \cite{johnson2016mimic, johnson2023mimic} and the eICU Collaborative Research Database \cite{pollard2018eicu} have enabled large-scale modeling over structured and semi-structured medical records, while i2b2/n2c2 shared tasks \cite{uzuner20112010} focus on extracting structured information from clinical text. In table reasoning, datasets such as TabFact \cite{chen2019tabfact} and FETAQA \cite{nan2022fetaqa} study fact verification and free-form question answering over tables, typically assuming that answers are grounded in explicit table entries. In contrast, our benchmark targets clinical-trial reasoning, where target attributes are often latent and must be inferred from visible row evidence through structured transformations.

\paragraph{Related Methods.}
Early table reasoning and semantic parsing approaches, such as Seq2SQL, TaBERT, and TAPAS \cite{zhong2017seq2sql, yin2020tabert, herzig2020tapas}, map natural language to executable queries or learned table representations and are strongest when the answer is explicitly represented in the schema. More recent LLM-based methods, including PAL and ReAct \cite{gao2023pal, yao2022react}, introduce intermediate reasoning and action steps, while TableGPT2, BlendSQL, and EHRAgent \cite{li2024table, glenn2024blendsql, shi2024ehragent} add stronger structure for reasoning over tables and clinical data. However, these methods either keep planning implicit within generation or rely on heavier agentic pipelines with iterative execution. In contrast, \textsc{SCoPE} explicitly decomposes the task into row selection, source-field grounding, and structured execution, making the reasoning process more transparent and controllable while achieving a better accuracy-efficiency tradeoff.

\section{Conclusion}

We introduced \textsc{SCoPE}, a planner-based framework for clinical trial table reasoning that separates row grounding, structured planning, \& row-level execution over visible evidence. Across same-backbone \& cross-model settings, explicit planning consistently improves grounded reasoning performance over direct prompting \& stronger structured baselines such as BlendSQL, EHRAgent, \& TableGPT2. The gains are especially pronounced for weaker implicit reasoners while remaining competitive for stronger backbones. These findings position clinical trial reasoning as a distinct table understanding problem \& suggest that lightweight planner-executor decomposition is an effective and cost-conscious approach for recovering latent clinical attributes from partially observed oncology tables.

\section*{Limitations}

Our study has several limitations. First, although we evaluate multiple strong open-weight and open-access backbones, we do not include the latest frontier proprietary models in the main experiments. It is therefore possible that stronger frontier systems would reduce some of the gains we observe explicit planning, or alternatively benefit even more from the planner-executor decomposition. Second, while we study same-backbone, cross-model, \& planner-coder ablations, we do not explore specialized fine-tuning of the planner \& executor into separate task-specific models. A natural extension would be to train custom planner models for structured reasoning and custom executor models for grounded row-level generation, which could further clarify how much of the observed gain comes from role specialization rather than prompting alone. Third, we do not perform a dedicated memorization study. Because the benchmark is derived from a compact clinical-trial table and programmatically expanded from a seed set, an additional analysis would be useful to determine whether some questions or transformations are partially recoverable from model memorization or prior exposure rather than purely from visible evidence. Finally, our benchmark is confined to a single clinical-trial table setting in oncology. Although this makes evaluation focused \& controlled, broader studies across additional clinical domains, schemas, \& larger real-world evidence tables would be needed to fully establish generality.

\section*{Ethics Statement}

This work is intended to support evidence review and clinical-trial exploration in oncology, not to guide patient-specific diagnosis, treatment selection, or other forms of clinical decision-making. \textsc{SCoPE} operates on a structured representation of public clinical-trial records and does not use patient-level notes, protected health information (PHI), or personally identifiable information (PII). The benchmark and evaluation tables are derived from publicly available trial data and contain only trial-level evidence. No human subjects were recruited for this study, and no new patient data were collected, accessed, or released.

Because large language models can produce incorrect, incomplete, or overconfident outputs, we design the system to remain auditable and grounded in the visible table evidence. The model is evaluated on row-aligned predictions derived from omitted fields, and the pipeline exposes intermediate planning and execution steps to reduce hidden reasoning and make failures easier to inspect. We restrict execution to read-only operations and apply schema validation and timeout safeguards to reduce the risk of unintended behavior. The benchmark construction procedure retains only read-only variants that execute successfully and return valid results, which improves consistency but may bias coverage toward more frequent query patterns.

The questions used to seed the benchmark were authored by domain experts to reflect realistic evidence-review needs in oncology and were reviewed to ensure that the expected answers are inferable from the table itself rather than from external medical knowledge. This design helps keep the task focused on grounded clinical table reasoning, but it also means that the benchmark does not capture the full complexity of real-world clinical information needs. In particular, the dataset is limited to a single oncology clinical-trial table and does not include broader clinical domains, longitudinal patient records, or free-text clinical narratives.

We note that model performance on this benchmark should not be interpreted as clinical reliability. In practice, outputs from \textsc{SCoPE} require expert review before use in any research or decision-making workflow. Future deployments should include access controls, logging, and clear user-facing warnings about model limitations and the possibility of hallucinated or unsupported outputs. The authors reviewed all technical claims and benchmark construction procedures, and any AI-assisted language editing used in manuscript preparation did not affect the scientific content or experimental results.

\section*{Acknowledgment}
This research was supported by the Mayo Clinic and Arizona State University Alliance for Health Care Collaborative Research Seed Grant Program (Award ID: AWD00041508; Sponsor Award ID: ARI-358187) for the project: `Artificial intelligence-assisted digital, living, interactive clinical practice guidelines for cancer providers and patients

\bibliography{custom}
\clearpage

\appendix

\section{Supplementary Material}
\label{app:supplementary}

\subsection{Planner Prompts}
\label{app:scope_prompts}

We provide the three inference-time prompt variants used by \textsc{SCoPE} below. Together, they implement the core stages of the pipeline: the selector-executor prompt identifies the candidate rows relevant to the question, the planner prompt converts the question and candidate table into an explicit reasoning plan, and the planner-executor prompt uses that plan to produce the final row-aligned predictions. Including these prompts makes clear how \textsc{SCoPE} externalizes row grounding, source-field selection, and answer generation rather than leaving them implicit in a single generation step.

\promptboxwide{prompts/planner_executor_executor.tex}{Plan Executor}{prompt:planner_executor}
\promptboxwide{prompts/planner_executor_planner.tex}{Planner}{prompt:planner}
\promptboxwide{prompts/planner_selector.tex}{Selector-Executor}{prompt:planner_selector}

\subsection{Prompting Baselines}
\label{app:prompting_baselines}

We provide the prompting templates used for the direct prompting baselines below. These are the prompts utilized by each of the direct prompting baselines that attempt to solve the reasoning questions in a single LLM call.

\promptboxwide{prompts/zeroshot.tex}{Zero-Shot Prompt}{prompt:zeroshot}
\promptboxwide{prompts/fs.tex}{Few-Shot Prompt}{prompt:fewshot}
\promptboxwide{prompts/cot.tex}{Chain-of-Thought Prompt}{prompt:cot}

\clearpage
\twocolumn[{
\subsection{Question Metadata}
\label{app:question_metadata}

\begin{center}
\includegraphics[width=0.9\textwidth]{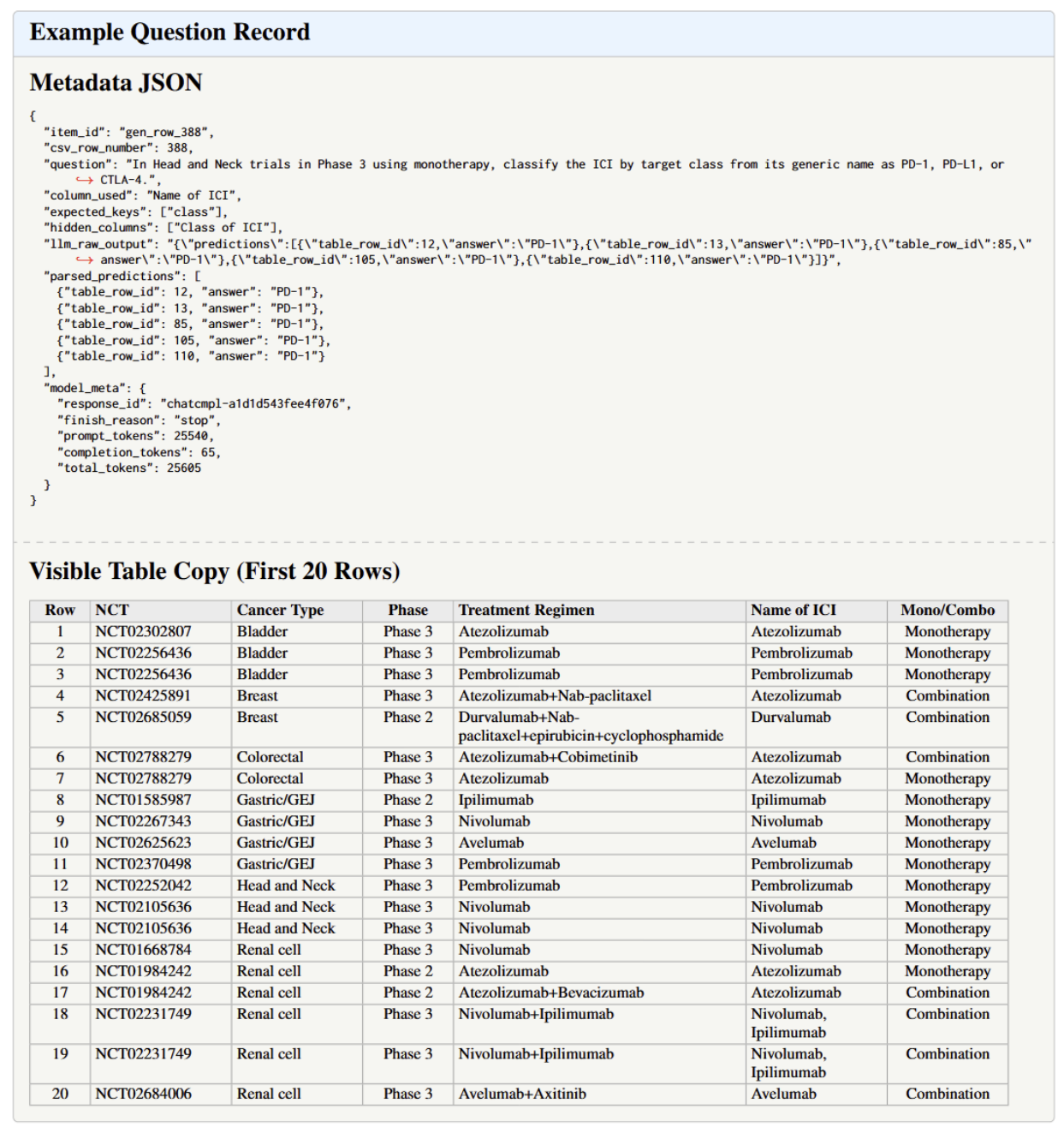}
\captionof{figure}{Example benchmark instance from the clinical trial reasoning dataset. The top panel shows the metadata JSON for question \texttt{gen\_row\_388}, including the natural-language question, the visible source column used for reasoning (\texttt{Name of ICI}), the hidden target column withheld at inference time (\texttt{Class of ICI}), the model's raw and parsed row-aligned predictions, and token-level generation metadata. The bottom panel shows the first 20 rows of the corresponding visible table copy provided to the model during inference. In this example, the task is to identify Head and Neck Phase 3 monotherapy trials and infer the immune checkpoint inhibitor target class (PD-1, PD-L1, or CTLA-4) from the generic ICI name}
\label{fig:question_metadata_example}
\end{center}

\vspace{0.5em}
}]

This section provides a representative example of the benchmark record format used in our experiments. It clarifies what information is exposed to the model at inference time, what target information is deliberately withheld for evaluation, and what prediction artifacts are stored for later analysis. More concretely, the example shows the natural-language question, the visible evidence-bearing column used for reasoning, the hidden target column omitted during inference, the model's raw response, the parsed row-aligned predictions used for scoring, and token-level generation metadata. The lower panel shows the corresponding visible table copy provided to the model. This makes clear that benchmark items are constructed to test grounded inference over visible evidence rather than direct retrieval from an explicit answer field. 

\twocolumn[{
\subsection{Error Analysis by Question Intent}
\label{app:error_analysis_intent}

\begin{center}
\small
\setlength{\tabcolsep}{6pt}
\renewcommand{\arraystretch}{1.14}
\begin{tabular*}{\textwidth}{@{\extracolsep{\fill}} p{3.7cm} p{8.3cm} c c c}
\toprule
\textbf{Intent Category} & \textbf{Typical Question Intent} & \textbf{N} & \textbf{$\Delta$F1} & \textbf{$\Delta$RJ} \\
\midrule
\textbf{Schema / structural labeling} &
Interpret compact schema-level fields or map them into standardized labels, such as PD-L1 requirement flags, number of arms, or clinical setting labels. &
168 & +4.93 & +7.94 \\

\textbf{Normalization / classification} &
Normalize visible values into canonical labels or booleans, such as control-arm type, combination type, or ICI class/brand mappings. &
341 & +1.36 & +2.30 \\

\textbf{Structured extraction} &
Extract agents, regimen names, chemotherapy backbones, or dosing schedules from treatment/control text. &
433 & +0.84 & +2.08 \\

\textbf{Endpoint reasoning} &
Determine endpoint role or related semantics from endpoint text, for example whether a listed endpoint is primary or secondary. &
181 & +0.45 & +1.21 \\

\textbf{Bucketing / discretization} &
Bucket continuous or ordinal values such as year, trial phase, sample size, or follow-up maturity into standardized categories. &
254 & -0.81 & +0.15 \\

\textbf{Metadata normalization / extraction} &
Recover or normalize metadata-derived values such as trial acronyms or PubMed-based identifiers. &
123 & -0.90 & -0.92 \\
\bottomrule
\end{tabular*}

\captionof{table}{Paired error-analysis summary by question intent for \textsc{SCoPE} (GPT-OSS executor, Llama-3.3 planner) versus the GPT-OSS chain-of-thought baseline. Questions are grouped heuristically by dominant intent. Positive $\Delta$ values indicate higher average per-question performance for \textsc{SCoPE}.}
\label{tab:error_analysis_by_intent}
\end{center}

\vspace{0.5em}
}]

We perform a paired error analysis against the GPT-OSS chain-of-thought baseline to understand where \textsc{SCoPE} helps and where it remains brittle. We compare the same 1{,}500 benchmark questions under two settings: (i) \textsc{SCoPE} with a GPT-OSS executor and a Llama-3.3 planner, and (ii) GPT-OSS CoT prompting. For each question, we compute $\Delta$F1 and $\Delta$RJ as the difference between \textsc{SCoPE} and the baseline, then group questions into broad intent categories using lightweight heuristics over the question templates. These categories are intended to capture the dominant reasoning demand of a question rather than define a formal ontology.

Several patterns emerge. \textsc{SCoPE} helps most on schema-centric questions, with the largest gains on \emph{Schema / structural labeling}, and it also improves \emph{Normalization / classification} and \emph{Structured extraction}. This suggests that explicit planning is especially useful when the task depends on identifying the correct evidence field and applying a constrained transformation over visible table content. \emph{Endpoint reasoning} also shows modest positive gains, indicating that planning helps with semantic disambiguation even when the source text remains noisy or underspecified.

The weakest categories are \emph{Bucketing / discretization} and \emph{Metadata normalization / extraction}. These questions often depend on exact thresholds, formatting conventions, acronym handling, or brittle string normalization, so errors can persist even when the correct rows and source fields are identified. Notably, the strongest categories show larger gains in $\Delta$RJ than in $\Delta$F1, suggesting that the main benefit of explicit planning is improved row-level grounding and alignment rather than only better aggregate answer recovery. Overall, the analysis indicates that \textsc{SCoPE} is most effective when the main challenge is schema grounding and transformation intent, and less effective when the remaining difficulty lies in exact normalization or surface-form matching.

\end{document}